\documentclass[11pt,a4paper]{article}
\usepackage[hyperref]{emnlp2020}
\usepackage{times}
\usepackage{latexsym}

\usepackage{microtype}
\usepackage{array}

\usepackage{graphicx}
\graphicspath{ {./} }
\aclfinalcopy 

\title{OpenFraming: We brought the ML; you bring the data. Interact with your data and discover its frames}

\author{
Alyssa Smith\thanks{ ~~These authors contributed equally.}
\And
David Assefa Tofu\footnotemark[1]
\And
Mona Jalal\footnotemark[1] \\
{asmithh@alum.mit.edu}, {\{\ davidat,\ jalal\ \}@bu.edu}, {edward.edberg.halim@gmail.com}, {ymsun@bu.edu}
\And
Edward E.~Halim
\And
Yimeng Sun
\AND 
Vidya Akavoor
\And
Margrit Betke
\And
Prakash Ishwar \\
{\{\ vidyaap,\ betke,\ pi,\ guolei,\ wijaya\ \}@bu.edu}
\And
Lei Guo
\And
Derry Wijaya
}

\date{}

\begin{document}
\maketitle
\begin{abstract} 
When journalists cover a news story, they can cover the story from multiple angles or perspectives. A news article written about COVID-19 for example, might focus on personal preventative actions such as mask-wearing, while another might focus on COVID-19's impact on the economy. These perspectives are called ``frames," which when used may influence public perception and opinion of the issue. We introduce a Web-based system for analyzing and classifying frames in text documents. 
 Our goal is to make effective tools for automatic frame discovery and labeling based on topic modeling and deep learning widely accessible to researchers from a diverse array of disciplines. To this end, we provide both state-of-the-art pre-trained frame classification models on various issues as well as a user-friendly pipeline for training novel classification models on user-provided corpora. Researchers can submit their documents and obtain frames of the documents. The degree of user involvement is flexible: they can run models that have been pre-trained on select issues; submit \textit{labeled} documents and train a new model for frame classification; or submit \textit{unlabeled} documents and obtain potential frames of the documents. The code making up our system is also open-sourced and well documented, making the system transparent and expandable. The system is available online at \url{http://www.openframing.org} and via our GitHub page \url{https://github.com/davidatbu/openFraming}.
 
\end{abstract}



\section{Introduction}
We live in a world saturated with media. Any major public issue, such as the ongoing COVID-19 pandemic and the Black Lives Matter protests, attracts tremendous attention from hundreds of thousands of news media outlets — traditional and emerging - around the world. The reporting angles on a single issue are often varied across different media outlets. In covering COVID-19, for example, some media outlets focus on government response and actions while others emphasize the economic consequences. Social science scholars call this process \emph{media framing}. To define, or to \emph{frame}, is ``to select some aspects of a perceived reality and make them more salient in a communicating text'' \cite{entman1993framing}. When used in news articles, frames can strongly impact public perception of the topics reported and lead to different assessments by readers \cite{hamborg2020media}, or even reinforce stereotypes and project explicit and implicit social and racial biases \cite{drakulich2015explicit,sap2019social}. 

Frame discovery in media text has been traditionally accomplished 
using methods such as quantitative content analysis \cite{krippendorff2018content}. However, in the emerging media environment, the sheer volume and velocity with which content is generated makes manual labeling increasingly intractable. To overcome this ``big data'' challenge, researchers have employed computational methods based on both unsupervised and supervised machine learning (ML) techniques. This has enabled users to detect frames automatically and robustly  \cite{akyurek-etal-2020-multi,liu2019detecting,tsur-etal-2015-frame}. 
However, these state-of-the-art computational tools are not readily accessible to social sciences scholars who typically do not have machine learning training. This hampers their ability to 
glean valuable insights from unprecedentedly large media datasets. 

Our goal is to make computational framing analysis accessible to researchers from a diverse array of disciplines. We present \href{http://www.openframing.org}{OpenFraming}, a user-friendly and interactive Web-based system that allows researchers to conduct computational framing analysis without having to write and debug complex code. There does, of course, exist click-and-run commercial software, but these tools often pose issues for researchers by their lack of transparency into their inner computational mechanisms. In contrast, our system is based on state-of-the-art research and our code is publicly available. While the focus of the project is on news media framing, the proposed system can also be used to implement other tasks such as sentiment detection or process other data types like social media data. 

Specifically, \href{http://www.openframing.org}{OpenFraming} can perform two types of computational framing analysis: 1) unsupervised topic modeling based on Latent Dirichlet Allocation (LDA; \citet{blei2003latent}), and 2) supervised learning using deep neural network Bidirectional Encoder Representations from Transformers (BERT; \citet{devlin2018bert}). Both approaches have been applied to media framing in communication research and are proven to be efficient and valid \cite{guo2016big,liu2019detecting}. 

When encountering a large set of unknown media data, researchers can employ the LDA-based approach to make sense of the data inductively \cite{guo2016big}. Using the LDA output, researchers can find the main threads of discourse in a corpus by examining the LDA ``topics'' associated with keywords that are most indicative of that particular thread of discourse. Ultimately, the ``topics'' can prove useful for frame discovery. However, the LDA output may not produce a useful framing model on its own. Because the method is unsupervised, the ``topics'' it creates may overlap with each other; appear to be irrelevant to the phenomenon being studied; or seem so ill-defined to the trained researcher that the results would not contribute to the framing literature. 

Therefore, our system also provides an alternative approach that allows domain experts (i.e., the users) to intervene in building the framing model. In this setting, the user can first employ the LDA-based approach to discover potential frames in the corpus. Then, using their domain-specific knowledge, they can manually label and upload a dataset to the system with frames suggested by the LDA model or uncovered from other explorations, whether machine-guided or not. We employ a BERT-based classification model to create a state-of-the-art frame classifier. Researchers can upload unlabeled documents 
to \href{http://www.openframing.org}{OpenFraming} and use the trained classifier to extract the frames.

To summarize, our system \href{http://www.openframing.org}{OpenFraming} has the following advantages: (1) It can process 
textual media data and detect frames automatically (2) It is accessible to researchers without computational backgrounds (3) It produces valid media frames based on peer-reviewed, state-of-the-art computational models (4) It provides many options for users to perform unsupervised ML, supervised ML, or both. In the supervised setting, the model trained on user-provided labeled data 
can be used to label a much larger dataset than would be feasible for human workers.

\section{Related Work}
A typical task in the field of communication research is the identification of topics, attributes, and frames in document collections to understand, for example, news media messages, elite discourse, and public opinion. Traditionally, scholars rely on content analysis approaches, both qualitative and quantitative, to manually annotate the data \cite{krippendorff2018content,lindlof2017qualitative}. In recent years, a group of communication researchers has taken advantage of advances in computational sciences and applied both unsupervised and supervised ML, to analyze large-scale communication text. In light of the growing importance of media and communication in our lives concerning agenda setting, framing, and biases, more and more computer scientists also joined this line of research and consider media framing to be a domain to apply their algorithms \cite{tsur-etal-2015-frame,field2018framing,liu2019detecting,akyurek-etal-2020-multi,hamborg2020media,sap2019social}.  


Within the world of unsupervised ML for text analysis, LDA-based topic modeling is one of the most widely used approaches in communication research (see \citet{maier2018applying} for a systematic review). The LDA algorithm generates a set number of ``topics'' associated with a list of terms. Researchers then review the terms and decide the label for each topic. Consider the news coverage of COVID-19 as an example. An LDA topic may include the terms \textit{pandemic}, \textit{job}, \textit{million}, \textit{economy}, and \textit{unemployment}, which can be labeled as the topic ``economic consequences''. Another topic may include the terms \textit{season}, \textit{player}, \textit{sport}, \textit{game}, and \textit{return}, and can be labeled as ``the impact on the sports industry''. \citet{guo2016big} made the first attempt to assess the efficacy and validity of the LDA-based approach in the context of journalism and mass communication research; furthermore, they prove that it is useful and efficient to obtain initial ideas about the data. 

Since a frame, explicitly defined, is ``a central organizing idea for news content 
that supplies a context and suggests what the issue is through the use of selection, emphasis, exclusion, and elaboration'' \cite{reese2001framing}, 
LDA-generated topics related to frames may elide the abstraction and nuance that the frames themselves contain. Framing scholars have identified a list of \textit{generic} and \textit{issue-specific} frames and argued that framing analysis should be built on the existing work to make a meaningful contribution to the literature \cite{guo2012transnational,nisbet2010knowledge,semetko2000framing}. 
This suggests that not all LDA-generated topics can be productively considered as frames. Using the running example of the COVID-19 coverage, while the LDA topic ``economic consequences'' corresponds to one of the generic frames identified earlier, it is debatable whether the topic discussing the impact on the sports industry can be interpreted as a frame. The LDA-based approach has other imperfections as well: it may generate meaningless ``topics'' or produce ``topics'' that contain unrelated or even conflicting information. Given this, the LDA approach is most useful for exploratory analysis. Although the LDA-generated topics are not necessarily equivalent to frames, the information can be used to infer potential frames 
for the next step of supervised frame analysis.

Unlike unsupervised ML, the supervised approach is a deductive research method and is used to identify pre-determined frames 
based on the literature. In communication research, scholars have used supervised ML algorithms such as support vector machines and deep learning models to identify frames in a media text. Two recent studies used BERT to identify frames in the news coverage of gun violence in the US; the studies both demonstrate a high level of accuracy \cite{akyurek-etal-2020-multi,liu2019detecting}.

The implementation of both unsupervised ML and supervised ML discussed above requires a computational background. Some social science scholars explore the methods themselves, and others choose to collaborate with colleagues in computer science. However, due to a lack of formal computer science training, it is often difficult for social science scholars to apply the computational models appropriately on their own. Also, not all scholars have the opportunities and resources for cross-disciplinary collaboration. Commercial software programs exist for this type of analysis, but most are costly and the algorithms they provide remain a black box. To overcome these challenges, we present \href{http://www.openframing.org}{OpenFraming}, a free and open-sourced Web-based system specialized in computational framing analysis. 

\section{System Architecture}
While we make a Web server that runs \href{http://www.openframing.org}{OpenFraming} publicly available, running one's copy of the system is 
also streamlined. This is possible through our release of a Docker container that orchestrates the various 
technologies used by our system. Concretely, this means that anyone ranging from the user who would like to have their own version
of the system on their personal computers, to bigger organizations who would like to host and extend the system
on more capable hardware, can get it up and running in minutes. The publicly available server, for example, 
was set up on an EC2 instance on Amazon Web Services (AWS) with minimal additional configuration.

The software that makes up the system includes Gensim's \cite{gensim2010}
Python interface to Mallet \cite{mallet2002} for LDA topic modeling; the transformers library for supervised classification \cite{wolf2019transformers}, 
Redis for queuing the jobs, SQLite for a database solution, Flask for the Web application backend, and jQuery and Bootstrap for the frontend.

\paragraph{Data Cleaning and Pre-processing for LDA}
While there is some flexibility regarding the format of the dataset (the system currently supports .xls, .xlsx, and .csv), it is nonetheless necessary that it at least contain a column labeled as ``Example''. This column will hold the text examples, with one document or, broadly speaking, textual entity, per row. 
LDA employs a bag-of-words model, where each document is understood as an unordered collection of words; to make the analysis more conducive to the discovery of useful topics, the system filters out extremely common and extremely rare words. The pre-processing steps we employ include the following: 
\begin{itemize}
 \item \textbf{Removing punctuation and digits.} this is a standard step in natural language processing (NLP) applications.
 \item \textbf{Removing stopwords}: stopwords are extremely common words, usually filtered out by default in NLP applications.
\item \textbf{Lemmatizing content}: this groups together different inflected forms of a word into a single entity.
\item \textbf{Setting minimum word length}: the system removes words shorter than 2 characters. 
\end{itemize}

\begin{figure}[ht!]
\includegraphics[width=8cm]{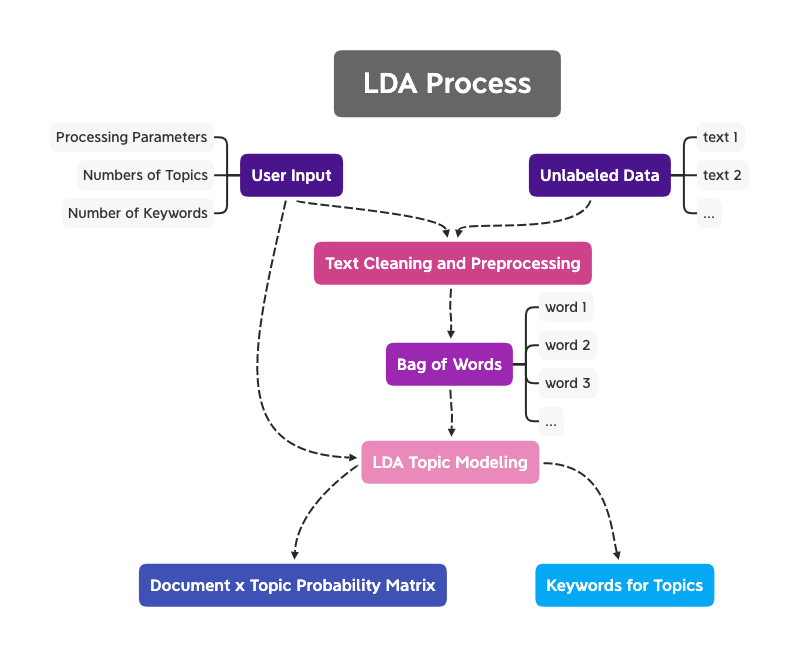}
\caption{LDA pipeline for topic Discovery}
\label{lda_pipeline}
\end{figure}
\paragraph{LDA for topic discovery}
The system runs LDA using the Mallet\cite{mallet2002} implementation and its preset parameter tuning. The random seed is set deterministically so that subsequent runs of the algorithm will yield the same results. LDA models each document as a probabilistic mixture of topics.
A topic is defined as a probability distribution over keywords. LDA iteratively updates the topic-keyword distributions to maximize the log-likelihood of the entire corpus. The system uses LDA to create a matrix mapping documents to weight vectors which quantify the contribution (weight) of each topic to the document; we can think of the weight vector as a probability distribution over topics for a particular document. Our system also produces a list of the most relevant keywords for each topic; the user can specify how many keywords they would like to be given before runtime. Because running LDA over a large corpus can be time-consuming, the user's part in monitoring the modeling finishes when they hit the ``submit'' button. The system then sends them an e-mail with a link to download the results of the analysis when it is ready. We also provide topic quality metrics, namely coherence, and perplexity, to aid researchers in refining the number of topics they choose to use to further analysis. Figure~\ref{lda_pipeline} provides a more detailed explanation of the LDA pipeline. 

\paragraph{Labeling Procedure For LDA Results}
When the LDA algorithm has completed, the user will receive its output, which contains a set of ``topics'', each of which is associated with a list of keywords. Communications researchers recommend that at least two researchers manually review the keywords and decide on a label for each topic. Ideally, for framing analysis, each label should correspond to one of the frames — generic or issue-specific — identified in the relevant literature. New labels may be created to signify topics or frames related to the specific issue. Given the limitations of the LDA approach, it is also possible that some ``topics'' may not be meaningful. 


\begin{figure*}[ht!]
\includegraphics[width=15cm]{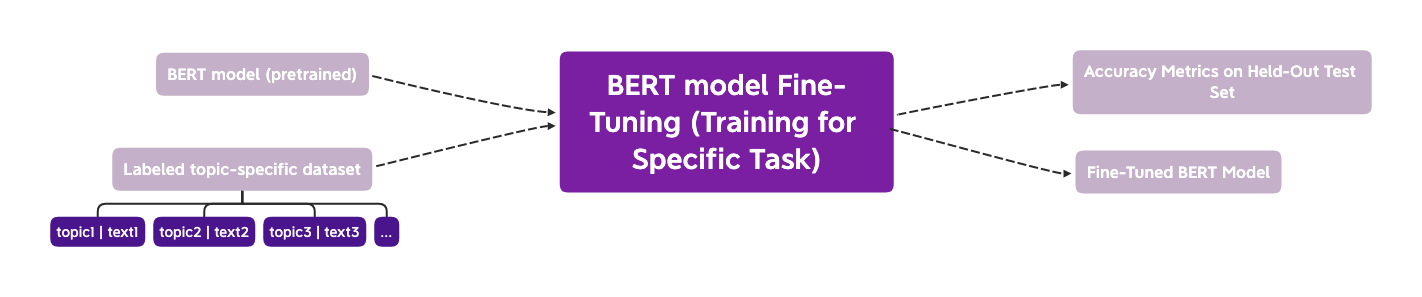}
\caption{BERT training/fine-tuning pipeline.}
\label{bert_pipeline}
\end{figure*}
\paragraph{Text classification using BERT}
BERT's masked language model \cite{devlin2018bert}, which builds on a deep Transformer's encoder architecture that relies on multi-layer self-attention to compute contextual representations of its input \cite{vaswani2017attention}, has shown 
impressive performance across a wide range of tasks, including framing analysis \cite{liu2019detecting}, when fine-tuned on labeled data for the task. 
However, there remains a significant access barrier for those with a non-computational
background to truly make use of BERT's wide-ranging applicability. To our knowledge, all publicly available 
Web services and software packages that make use of BERT either constrain the end-user to one specific fine-tuned model
(for example, fine-tuned on a specific sentiment analysis dataset), or, they require their users to 
be prepared to write code to fine-tune and further predict on a custom dataset. \href{http://www.openframing.org}{OpenFraming} makes it possible for those without a computational background to take advantage of BERT's impressive fine-tuned performance
on a custom dataset of their own.

When the user uploads \textit{labeled} data for training and testing or \textit{unlabeled} data for inference, our system either fine-tunes a new BERT model or uses our existing fine-tuned BERT for classifying the frame labels in the data. For fine-tuning, our system uses the standard configuration of BERT's internal architecture, and uses one set of training parameters recommended for BERT: a learning rate of 5e-5, 3 epochs of fine-tuning training,
and a batch size of 8. 

Once training or inference are completed, the user receives an e-mail with a download link to the frame prediction results on their data. 
In the case of fine-tuning a new BERT model on user-provided labeled data, we also provide accuracy on user-provided test data and the newly fine-tuned BERT model that the user can download. Figure~\ref{bert_pipeline} provides a more detailed description of the BERT training pipeline.

\paragraph{Labeling Procedure for Training a New BERT Model}
With the feature ``Training BERT – Do-it-yourself method,'' users can train a new BERT classification model using their own labeled data. In social science research, quantitative content analysis is one of the most widely used methods for labeling visual and textual content \cite{kripendorff2004content, riffe2019analyzing}. The approach involves drawing a representative sample of data; training two or more human coders on a labeling protocol to identify patterns in content, and measuring intercoder reliability between their 
coding results. Once the coders reach a certain degree of intercoder reliability, they can start labeling the remaining data independently. Communications researchers have recently suggested that crowdsourcing, if appropriately implemented, can be a valid alternative to annotating media messages \cite{guo2019accurate,lind2017content}. 
The labeled data can then be uploaded to our system to train a new BERT model. 

\paragraph{Available Pre-Trained BERT Models for Frame Classification}
For the feature ``Using BERT – off-the-shelf classifier,'' users can use models that we have fine-tuned on benchmark frame datasets to classify their unlabeled data. We make available models that can label frames on issues that include (1) immigration, (2) tobacco-use, (3) same-sex marriage (fine-tuned on Media Frame Corpus dataset \cite{card2015media}), (4) US Gun Violence issue (fine-tuned on Gun Violence Frame Corpus \cite{liu2019detecting}), or (5) COVID-19. To validate the performance of our fine-tuned model and the quality of its predictions, users can label a sample of their documents using the aforementioned approaches — quantitative content analysis and crowdsourcing — and compare the manual and machine-generated labels. 


\section{User Interface and Site Design}

\begin{figure}[ht!]
\includegraphics[width=\linewidth]{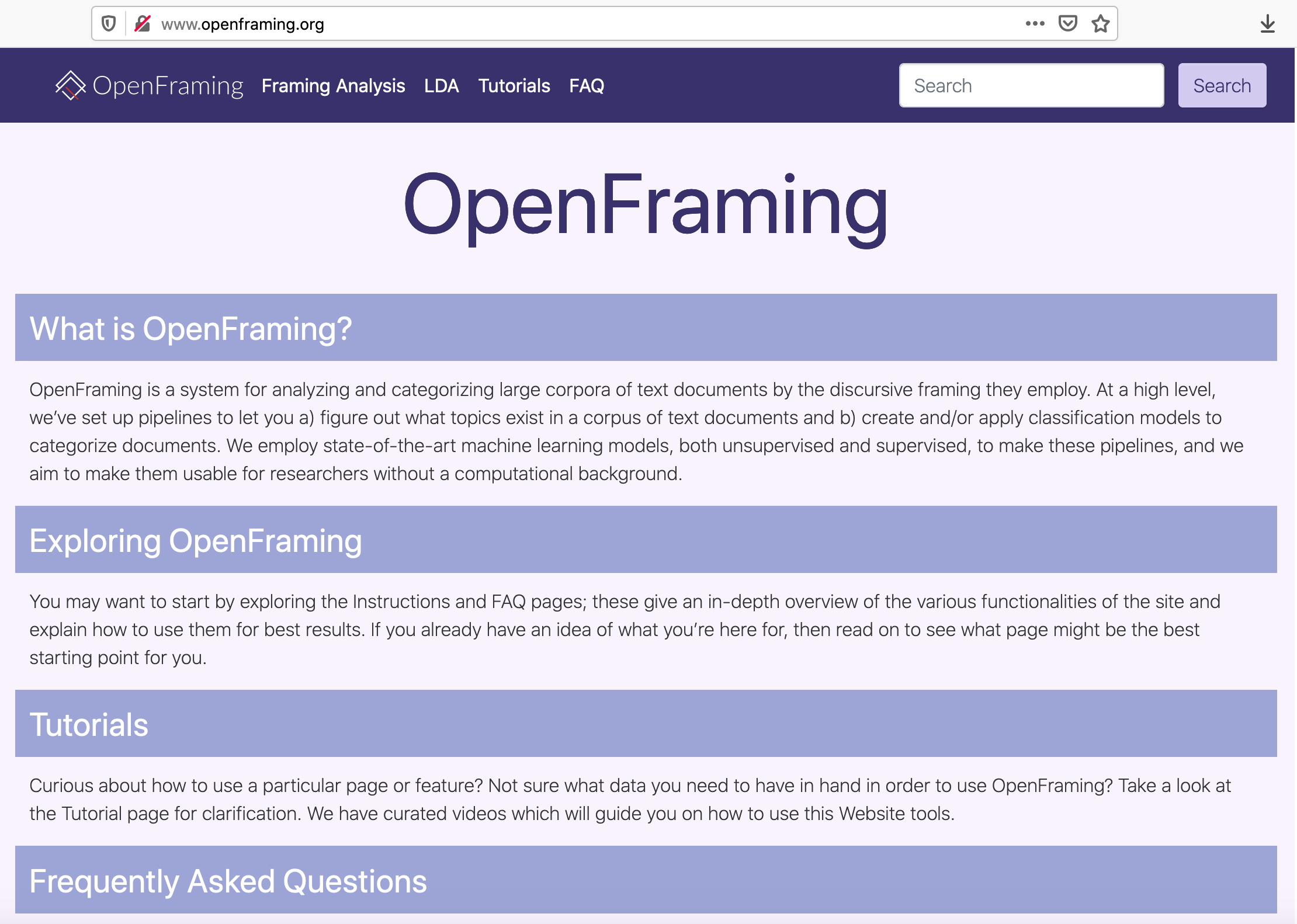}
\caption{The landing page of  \href{http://www.openframing.org}{openframing.org}}
\label{landing_page}
\end{figure}

 Our demo Website includes framing analysis as well as LDA topic discovery utilities. Additionally, our landing page provides an introduction to the user explaining what various building blocks of our Website are (Figure~\ref{landing_page})
 
\begin{figure}[ht!]
\includegraphics[width=\linewidth]{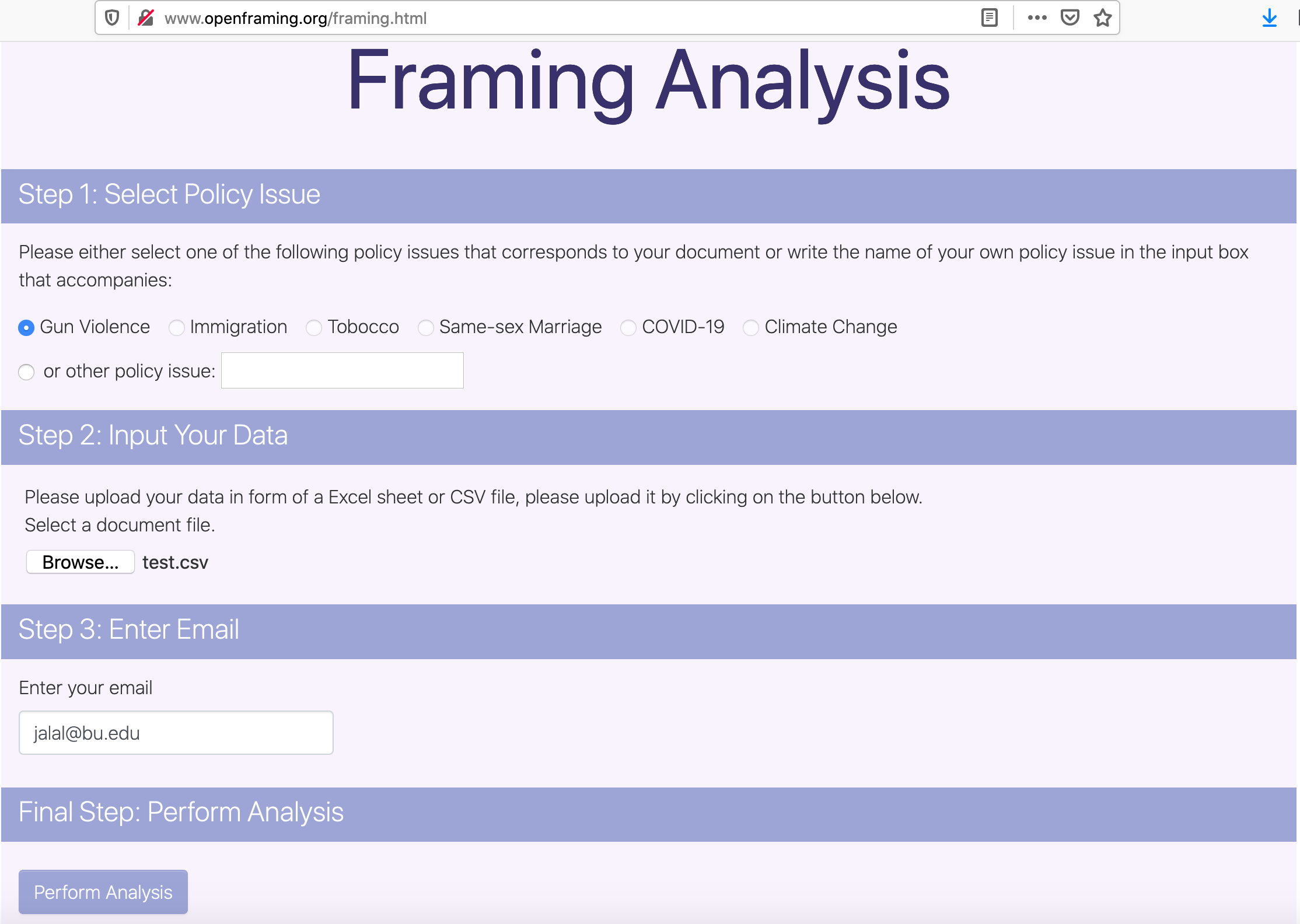}
\caption{Framing analysis Web page}
\label{Figure2}
\end{figure}

 Our framing analysis page (Figure~\ref{Figure2}) is created to accommodate two use cases. Either the user inputs a file for framing classification and chooses one of the policy issues for which we already have pre-trained models (e.g. \textit{Immigration}), or picks one of the policy issues of their choosing (e.g. \textit{Labor Market Inequality}). If the user chooses their own policy issue for which we don't have a pre-trained model, they are required to also upload a sizable dataset labeled with frames (containing approximately 100 documents for each frame) 
 so that the system can train a new BERT-based framing classifier for the issue in the backend.

\begin{figure}[ht!]
\includegraphics[width=\linewidth]{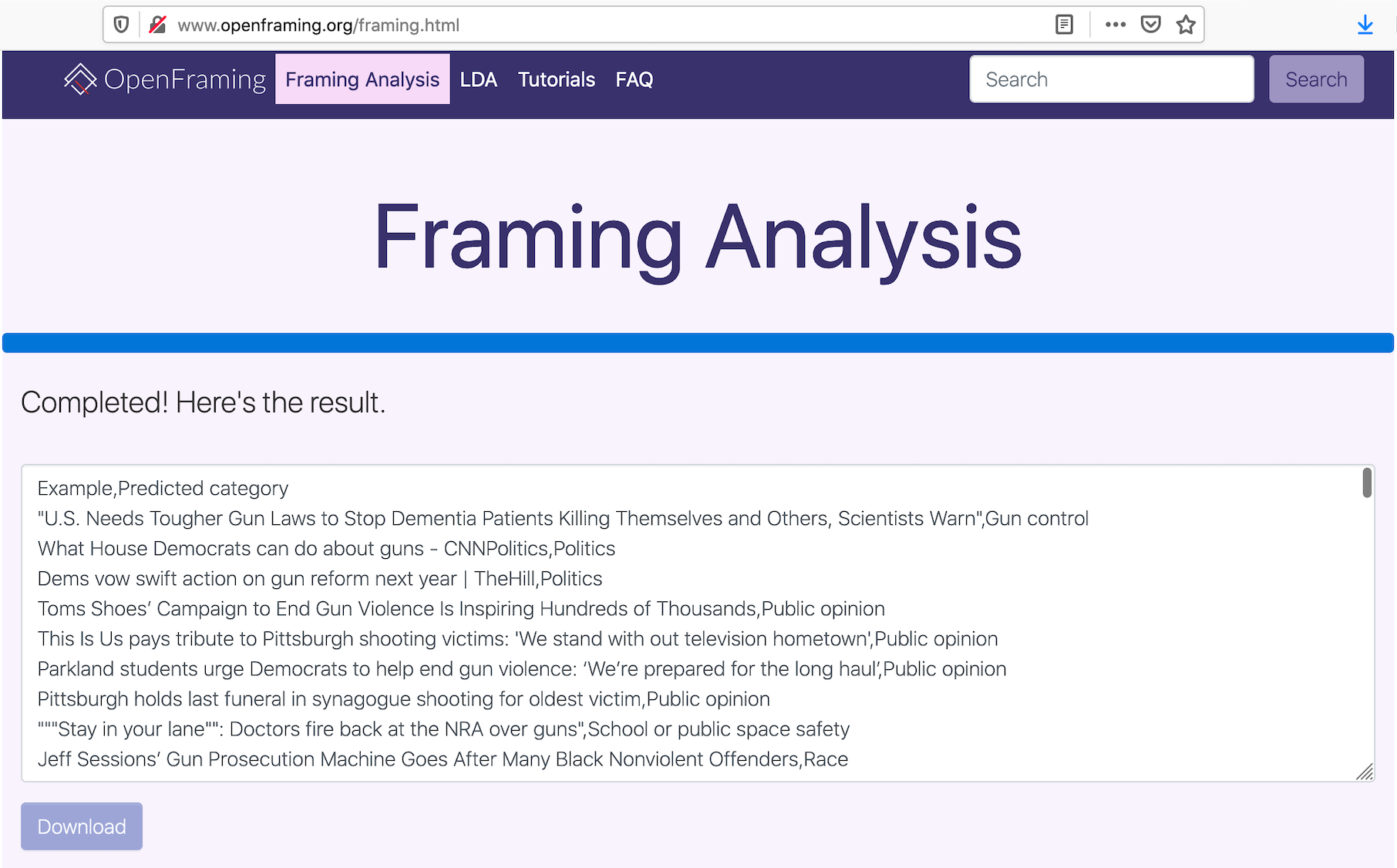}
\caption{A snapshot of framing classification results}
\label{framing_UI}
\end{figure}

Once the backend has completed running inference on the pre-defined and pre-trained policy issues or completed the training and inference on user-defined policy issue, the results will be shown dynamically on the same page (Figure~\ref{framing_UI}). The user can then scroll through the predicted results and download the results to their local machines.

\begin{figure}[ht!]
\includegraphics[width=\linewidth]{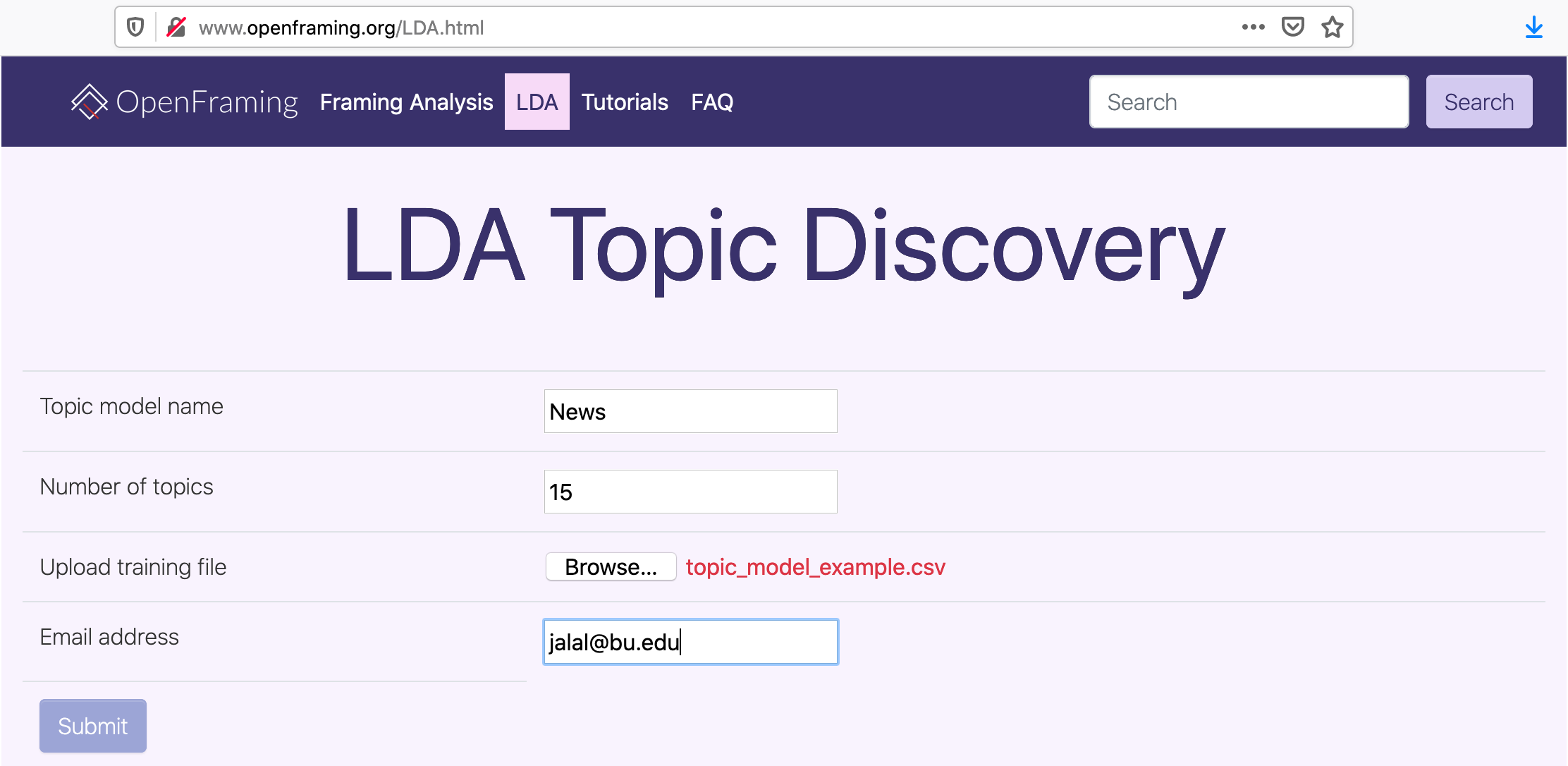}
\caption{LDA topic discovery page}
\label{LDA_UI}
\end{figure}

\begin{figure}[ht!]
\includegraphics[width=\linewidth]{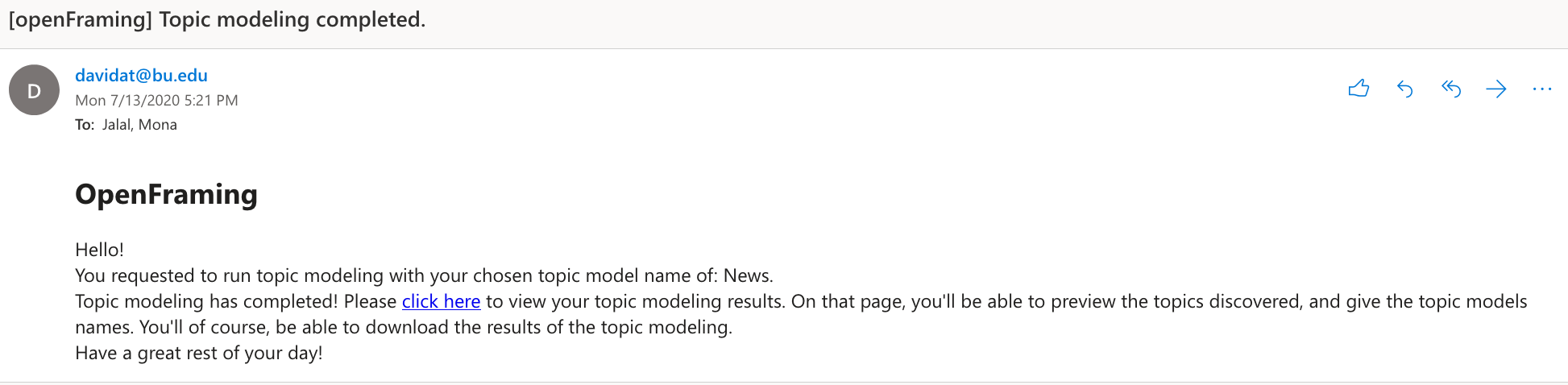}
\caption{LDA results are ready and e-mailed to the user.}
\label{LDA_results_email}
\end{figure}

\begin{figure}[ht!]
\includegraphics[width=\linewidth]{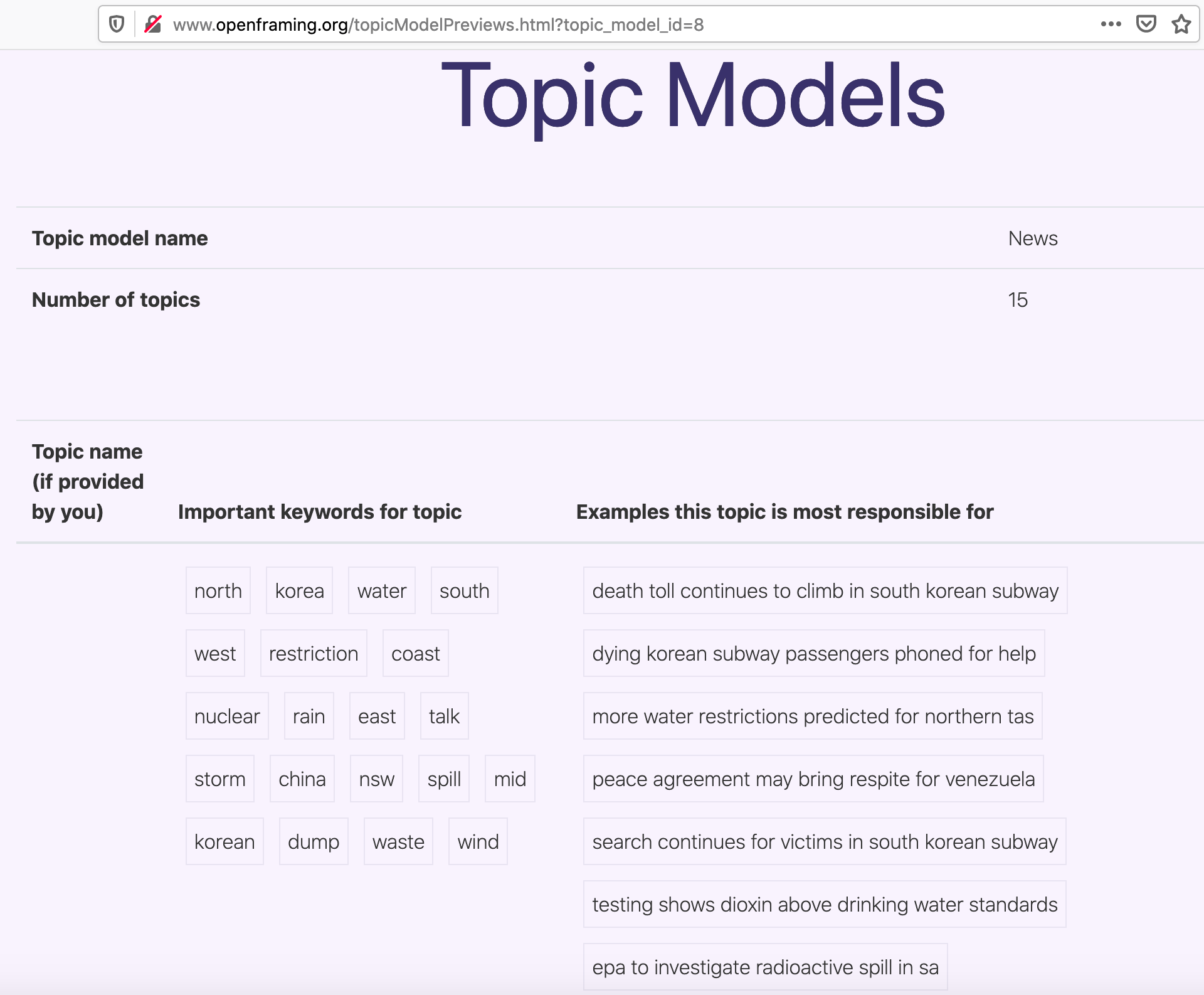}
\caption{A snapshot of one of the topics discovered by LDA on `A Million News Headline' dataset, the keywords for the topic, and the headlines labeled with the topic. 
}
\label{LDA_results}
\end{figure}

Here, we illustrate the topic discovery functionality of \href{http://www.openframing.org}{OpenFraming} (Figure~\ref{LDA_UI}) using a sample from the Kaggle `A Million News Headline' dataset\footnote{\href{https://www.kaggle.com/therohk/million-headlines}{https://www.kaggle.com/therohk/million-headlines}}. Once topics are discovered, we send the topics and their keywords as well as the document topic probabilities to user's provided e-mail (Figure~\ref{LDA_results_email} and Figure~\ref{LDA_results}). 

We have also created a screencast video demonstrating the use of the system, which can be accessed at \url{https://www.youtube.com/watch?v=u8SJAZ-EbgU}. 

\section{Conclusion and Future Work}
We have introduced OpenFraming, a Web-based system for analyzing and classifying frames in the text documents. OpenFraming is designed to lower the barriers to applying machine learning for frame analysis, including giving researchers the capability to build models using their own labeled data. Its architecture is designed to be user-friendly and easily navigable, empowering researchers to comfortably make sense of their text corpora without specific machine learning knowledge.

In future work, we hope to incorporate semi-supervised machine learning methods to allow researchers to iterate quickly on models; if a researcher submits a dataset with a relatively small number of labels, for example, the system will eventually be able to generate labels for the much larger unlabeled dataset, creating a synthetic training set for the BERT supervised model to train on. 

\bibliographystyle{acl_natbib}
\bibliography{anthology,emnlp2020}
\end{document}